\documentclass[sigconf]{acmart}

\AtBeginDocument{%
  \providecommand\BibTeX{{%
    \normalfont B\kern-0.5em{\scshape i\kern-0.25em b}\kern-0.8em\TeX}}}

\setcopyright{acmcopyright}
\copyrightyear{2023}
\acmYear{2023}
\acmDOI{XXXXXXX.XXXXXXX}

\acmConference[Conference acronym 'XX]{Make sure to enter the correct
  conference title from your rights confirmation emai}{2023}{Woodstock, NY}
\acmPrice{15.00}
\acmISBN{978-1-4503-XXXX-X/18/06}

\usepackage{float}
\usepackage[utf8]{inputenc}
\usepackage{amsmath}

\usepackage{amssymb}
\usepackage[capitalise]{cleveref}
\usepackage{natbib}
\usepackage{multirow}

\DeclareMathOperator*{\E}{\mathbb{E}}

\DeclareMathOperator*{\MI}{\mathcal{I}}
\DeclareMathOperator*{\MU}{\mathcal{U}}

\newcommand{\x}[1]{{\color{red}\bf{}}}

\newsavebox{\boxone}
\newsavebox{\boxtwo}
\newsavebox{\boxthree}

\newlength{\narrow}
\newlength{\cnarrow}
\newcommand{\topline}{
  \hrule
  \vskip .5\baselineskip}
\newcommand{\bottomline}{
  \vskip 2pt
  \hrule}

\newcommand{\chbox}[2]{
  \hbox to #1{\hss\vtop{#2}\hss}}

\newcommand{\nchbox}[1]{
  \chbox{\narrow}{#1}}

\newcommand{\cnchbox}[1]{
  \chbox{\cnarrow}{#1}}

\newcommand{\fcode}[1]{
  
  \chbox{\textwidth}{\tgrind\input{#1}}}

\newcommand{\ncode}[1]{
  
  \chbox{\narrow}{\tgrind\input{#1}}}

\def\nfig#1#2#3{
  \vtop{\nchbox{#1}
  \hbox to\narrow{\parbox{\narrow}{\caption{#2}\label{#3}}}}}

\newcommand{\cncode}[1]{
  \chbox{\cnarrow}{\tgrind\input{#1}}}

\def\codefiggen[#1]#2#3#4#5#6{
  \begin{figure}[#1]
  #5
  \fcode{#2}
  \center\parbox{.9\textwidth}{\caption{#3}\label{#4}}
  #6
  \end{figure}}

\def\codefig[#1]#2#3#4{
  \codefiggen[#1]{#2}{#3}{#4}{}{}}

\def\codefigline[#1]#2#3#4{
  \codefiggen[#1]{#2}{#3}{#4}{\topline}{\bottomline}}

\def\doublefiggen[#1]#2#3#4#5#6#7#8#9{
  \begin{figure}[#1]
  #8
  \hbox to \textwidth{
  \nfig{#2}{#3}{#4}
  \hfil
  \nfig{#5}{#6}{#7}}
  #9
  \end{figure}}

\def\doublefig[#1]#2#3#4#5#6#7{
  \doublefiggen[#1]{#2}{#3}{#4}{#5}{#6}{#7}{}{}}

\def\doublefigline[#1]#2#3#4#5#6#7{
  \doublefiggen[#1]{#2}{#3}{#4}{#5}{#6}{#7}{\topline}{\bottomline}}

\def\doublecodefig[#1]#2#3#4#5#6#7{
  \doublefig[#1]{\ncode{#2}}{#3}{#4}{\ncode{#5}}{#6}{#7}}

\def\doublecodefigline[#1]#2#3#4#5#6#7{
  \doublefigline[#1]{\ncode{#2}}{#3}{#4}{\ncode{#5}}{#6}{#7}}

\newcommand{\codepair}[4]{\vbox{
  \hbox{\ncode{#1} \hfil \ncode{#3}}
  \vskip .3\baselineskip plus .3\baselineskip
  \hbox{\hbox to\narrow{#2\hfil} \hfil \hbox to\narrow{#4\hfil}}}}

\def\codepairfig[#1]#2#3#4#5#6#7{
  \begin{figure}[#1]
  \codepair{#2}{#3}{#4}{#5}
  \center\parbox{.9\textwidth}{\caption{#6}}
  \label{#7}
  \end{figure}}

\def\cncodepairfiggen[#1]#2#3#4#5#6#7{
  \begin{figure}[#1]
  #6
  \hbox{\cncode{#2}\hfil\cncode{#3}}
  \center\parbox{.9\columnwidth}{\caption{#4}\label{#5}}
  #7
  \end{figure}}

\def\cncodepairfig[#1]#2#3#4#5{
  \cncodepairfiggen[#1]{#2}{#3}{#4}{#5}{}{}}

\def\cncodepairfigline[#1]#2#3#4#5{
  \cncodepairfiggen[#1]{#2}{#3}{#4}{#5}{\topline}{\bottomline}}

\def\doublefigOnecap*[#1]#2#3#4#5{
  \begin{figure*}[#1]
  \hbox to \textwidth{
  \nchbox{#2}
  \hfil
  \nchbox{#3}}
  \caption{#4}
  \label{#5}
  \end{figure*}}

\def\doublefigOnecap[#1]#2#3#4#5{
  \begin{figure}[#1]
  \topline
  \hbox to \columnwidth{
  \cnchbox{#2}
  \hfil
  \cnchbox{#3}}
  \caption{#4}
  \label{#5}
  \bottomline
  \end{figure}}

\def\PSfig[#1]#2#3#4{
 \begin{figure}
 \centerline{\psfig{file=#2,width=\columnwidth}}
 \caption{{#3}}
 \label{#4}
 \end{figure}}

\def\PSfiglines[#1]#2#3#4{
 \begin{figure}[#1]
 \topline
 \centerline{\psfig{file=#2,width=\columnwidth}}
 \caption{{#3}}
 \label{#4}
 \bottomline
 \end{figure}}

\def\PSfiglinesht[#1]#2#3#4#5{
 \begin{figure}[#1]
 \topline
 \centerline{\psfig{file=#2,height=#3}}
 \caption{{#4}}
 \label{#5}
 \bottomline
 \end{figure}}

\def\doublePSfig[#1]#2#3#4#5#6{
  \doublefigOnecap[#1]
    {\cnchbox{\psfig{file=#2,height=#4}}}
    {\cnchbox{\psfig{file=#3,height=#4}}}
    {#5}
    {#6}}

\newlength{\boxwidth}
\newcommand{\bproof}{{\bf Proof Sketch: }}
\newcommand{\eproof}{\mbox{$\Box$}}

\def\tabdoublecode#1#2#3#4{
 \begin{figure*}[t]
 \topline\vs{-.4}
 \hbox to \columnwidth{
 \vtop{\small
 \begin{tabbing}
 #1
 \end{tabbing}}
 \hfil
 \hfil
 \hfil
 \vtop{\small
 \begin{tabbing}
 #2
 \end{tabbing}}
 }
 \caption{#3\label{#4}}
 \bottomline
 \end{figure*}
}
\def\tabtriplecode#1#2#3#4#5{
 \begin{figure}
 \topline\vs{-.4}
 \hbox to \columnwidth{
 \vtop{\small
 \begin{tabbing}
 #1
 \end{tabbing}}
 \hfil
 \hfil
 \hfil
 \vtop{\small
 \begin{tabbing}
 #2
 \end{tabbing}}
 \hfil
 \hfil
 \hfil
 \vtop{\small
 \begin{tabbing}
 #3
 \end{tabbing}}
 }
 \caption{#4\label{#5}}
 \bottomline
 \end{figure}
}

\newtheorem{lemma}{Lemma}
\newcommand{\blemma}{\begin{lemma}}
\newcommand{\elemma}{\end{lemma}}

\newtheorem{thm}{Theorem}
\newcommand{\bthm}{\begin{thm}}
\newcommand{\ethm}{\end{thm}}

\newtheorem{defin}{Definition}
\newcommand{\bdefin}{\begin{defin}}
\newcommand{\edefin}{\end{defin}}

\newtheorem{observation}{Observation}
\newcommand{\bobserv}{\begin{observation}}
\newcommand{\eobserv}{\end{observation}}

\newtheorem{coroll}{Corollary}
\newcommand{\bcoroll}{\begin{coroll}}
\newcommand{\ecoroll}{\end{coroll}}

\newcommand{\vs}[1]{\vspace{#1cm}}
\newcommand{\be}{\begin{equation}}
\newcommand{\ee}{\end{equation}}

\newcommand{\bdesc}{\begin{description}}
\newcommand{\edesc}{\end{description}}
\newcommand{\benum}{\begin{enumerate}}
\newcommand{\eenum}{\end{enumerate}}
\newcommand{\bitem}{\begin{itemize}}
\newcommand{\eitem}{\end{itemize}}
\newcommand{\bcenter}{\begin{center}}
\newcommand{\ecenter}{\end{center}}
\newcommand{\btabular}{\begin{tabular}}
\newcommand{\etabular}{\end{tabular}}
\newcommand{\beqnarr}{
 \begin{eqnarray}}
\newcommand{\eeqnarr}{\end{eqnarray}}

\begin{document} 
\title{(Debiased) Contrastive Learning Loss for Recommendation (Technical Report)}

\author{Ruoming Jin}
\email{rjin1@kent.edu}
\author{Dong Li}
\email{dli12@kent.edu}
\affiliation{%
  \institution{Kent State University}
  \city{Kent}
  \state{Ohio}
  \country{USA}
  \postcode{43017-6221}
}

\begin{abstract}
In this paper \footnote{{This manuscript was initially reviewed at a conference in February 2023.}}, we perform a systemic examination of the recommendation losses, including listwise (softmax), pairwise(BPR), and pointwise (mean-squared error, MSE, and Cosine Contrastive Loss, CCL)  losses through the lens of contrastive learning. 
We introduce and study both debiased InfoNCE and mutual information neural estimator (MINE), for the first time, under the recommendation setting. 
We also relate and differentiate these two losses with the BPR loss through the lower bound analysis. 
Furthermore, we present the debiased pointwise loss (for both MSE and CCL) and theoretically  certify both iALS and EASE, two of the most popular linear models, are inherently debiased. 
The empirical experimental results demonstrate the effectiveness of the debiased losses and newly introduced mutual-information losses outperform the existing (biased) ones. 
\end{abstract}

\maketitle
\section{Introduction}

Machine learning, including recommender systems \cite{charubook,li2023impressioninformed}, has played a pivotal role in transforming the way humans interact with and explore information on the internet \cite{cellpad}, which has been ongoing for several years and continues to evolve with the integration of AI-generated content \cite{li2022generationaugmented}, significantly impacting the lives of individuals \cite{bot}. As recommendation algorithms continue to boom \cite{charubook,zhang2019deep}, largely thanks to deep learning models, many models' real effectiveness and claimed superiority has been called into question, such as various deep learning models vs. simple linear and heuristic methods ~\cite{RecSys19Evaluation}, and the potential  ``inconsistent'' issue of  using item-sampling based evaluation \citep{KricheneR22,li2020sampling,Jin@AAAI21,dong@2023aaai,li@acm23}, among others. A major problem in existing recommendation research is their ``hyper-focus'' on evaluation metrics with often weak or not-fully-tuned baselines to demonstrate the progress of their newly proposed model ~\cite{RecSys19Evaluation}. However, recent research has shown that properly evaluating (simple) baselines is indeed not an easy task~\cite{difficulty@rendle}. 

 To deal with the aforementioned challenges, aside from standardized benchmarks and evaluation criteria, research communities have devoted efforts to developing and optimizing simple and stronger baseline models, such as the latest effort on fine-tuning iALS~\cite{ials_revisiting,ials++} and SimpleX (CCL)~\cite{simplex}. In the meantime, researchers ~\cite{jin@linear} have started to theoretically analyze and compare different models, such as matrix factorization (iALS) vs. low-rank regression (EASE). By simplifying and unifying different models under a similar closed form, they observe how different approaches ``shrink'' singular values differently and thus provide a deep understanding and explain how and why the models behave differently, and the exact factor leads to a performance gap. However, the above work is limited only to linear models, and their regularization landscapes ~\cite{li2022on}. 

Taking the cue from the aforementioned research, this paper aims to  rigorously analyze and compare the loss functions,  which play a key role in designing recommendation models~\cite{simplex}. This paper is also particularly driven by the emergence of {\em contrastive learning} loss~\cite{SimCLR,infonce,debiased} has shown great success in various learning tasks, such as computer vision and NLP, etc.  Even though there are a few research works ~\cite{cl-rec,alignment-uniformity} in adopting contrastive learning paradigms into individual recommendation models, a systematic understanding of the recommendation loss function from the lens of contrastive learning is still lacking. For instance, how recommendation models can take advantage of (debiased) contrastive loss~\cite{debiased} (including mutual information-based losses \citep{MINE}? How pairwise BPR~\cite{bpr} (Bayesian Pairwise Ranking) loss relate to these recent contrastive losses~\cite{debiased,infonce,MINE}?  How can we debias the $L2$ (mean-squared-error, MSE) or $L1$ (mean-absolute-error, MAE) loss function with respect to the contrastive learning loss? Following this, does the well-known linear models, such as iALS \citep{ials_revisiting,hu2008collaborative,ials++} and EASE \citep{ease}, needs to be debiased with respect to contrastive learning losses?

To answer these questions, we made the following contribution to this paper: 
\begin{itemize}
\item (\cref{sec:softmax}) We revisit the softmax loss (and BPR loss) using the latest debiased contrastive loss~\cite{debiased}, which provides a natural debias approach instead of typically rather sophisticated debias methods through propensity score ~\cite{unbiased-propensity}. We also introduce a mutual information neural estimator (MINE) loss~\cite{MINE} to the recommendation setting. To the best of our knowledge, this is the first study to study MINE for recommendation models.  Experimental results demonstrate its (surprising) performance compared with other known losses.  Finally, through the lower bound analysis, we are able to relate the contrastive learning losses and BPR loss. 
\item (\cref{sec:linear}) We generalize the debiased contrastive loss ~\cite{debiased} to  the pointwise loss (MSE ~\cite{hu2008collaborative} and CCL~\cite{simplex}) in recommendation models and design the new debiased pointwise loss. We then examine the well-known linear models with pointwise loss, including iALS and EASE; rather surprisingly, our theoretical analysis reveals that both are inherently debiased under the reasonable assumption with respect to the contrastive learning losses.    
\item (\cref{sec:experiment}) We experimentally validate the debiased InfoNCE and point-wise losses indeed perform better than the biased (commonly used) alternatives. We also show the surprising performance of the newly introduced mutual information-based recommendation loss and its refinement (MINE and MINE+). 
 \end{itemize}
 Finally, Section ~\ref{sec:related} reviews the related work, and Section ~\ref{sec:conclusion} concludes. This paper towards to build a loss function theory for a recommendation based on contrastive learning.

\section{Contrastive Recommendation Loss} 
\label{sec:softmax}
 In this section, we introduce Debiased InfoNCE~\cite{infonce} and Mutual-Information based Loss (MINE) ~\cite{MINE} to the recommendation settings. Note that, the study in this section, is not to claim those losses are new (they stem from the latest contrastive learning research~\cite{unbiased,debiased}), but to bring them into the recommendation community. Indeed, to our surprise, those losses from contrastive learning have not been fully explored and leveraged by the recommendation models. To reemphasize, our goals are two folds in this section: 1) demonstrating  the benefits of these contrastive learning inspired losses are useful for recommendation models; 2) utilizing the lens of contrastive learning to carefully reexamine the losses recommendation community have designed before towards better understanding and unifying them. 

\noindent{\bf Notation:} 
We use $\mathcal{U}$ and $\mathcal{I}$ to denote the user and item sets, respectively. For an individual user $u\in \mathcal{U}$, we use $\mathcal{I}^+_u$ to denote the set of items that have been interacted (such as Clicked/Viewed or Purchased) by the user $u$, and $\mathcal{I}\backslash \MI^+_u$ to represent the remaining set of items that have not been interacted by user $u$. Similarly, $\MU^+_i$ consists of users interacting with item $i$, and $\MU\backslash \MU_i^+$ includes remaining users.
Often, we denoted $r_{ui}=1$ if item $u$ is known to interact with of item $i$ and $r_{ui}=0$ if such interaction is unknown.

Given this, most of the recommendation systems utilize either matrix factorization (shallow encoder) or deep neural architecture to produce a latent user vector $v_u$ and a latent item vector $v_i$, then use their inner product ($<v_u, v_i>$) or cosine ($<v_u/||v_u||, v_i/||v_i||>$) to produce a similarity measure $\hat{y}_{ui}$. The loss function is then used to produce a (differentiable) measure to quantify and encapsulate all such $\hat{y}_{ui}$. 

\subsection{Basic (Sampled) Softmax and InfoNCE loss}
\noindent{\bf Softmax Loss:} 
In recommendation models, a softmax function is often utilized to transform the similarity measure into a probability measure~\cite{youtube,cl-rec}: $
p(i|u) = \frac{\exp(\hat{y}_{ui})}{\sum_{j \in \mathcal{I}} \exp(\hat{y}_{uj})}$. Given this, the maximal likelihood estimation (MLE) can be utilized to model the fit of the data through the likelihood 
($\Pi_{u\in \mathcal{U}, i \in \MI^+_u} p(i|u)$), and the negative log-likelihood serves as the loss function: 
\begin{equation} 
\mathcal{L}_{soft} = - \E_u  \log \sum_{i \in \mathcal \MI^+_u } \frac{exp(\hat{y}_{ui})}{\sum_{j \in \mathcal{I}} exp(\hat{y}_{uj})}
\end{equation} 

Note that in the earlier research, the sigmoid function has also adopted for transforming similarity measure to probability: $p(i|u) =\sigma(\hat{y}_{ui})$. However, it has shown to be less effective ~\cite{ncfmf}. Furthermore, the loss function is a binary cross-entropy where the negative sample part is eliminated as being treated as $0$. Finally, the main challenge here is that the denominator in the loss sums over all possible items in $\mathcal{I}$, which is often impossible in practice and thus requires approximation.  In the past, the sampled softmax approaches typically utilized the important sampling for estimation~\cite{sample-softmax}: 
\begin{equation}
    \mathcal{L}_{ssoft}= - \E_u \log \sum_{i \in \mathcal \MI^+_u } \frac{exp(\hat{y}_{ui})}{exp(\hat{y}_{ui}) + \sum_{j=1; j \sim p^-_u}^N  exp(\hat{y}_{uj})/p_u^-(j)}
\end{equation}
where, $p_u^-$ is a predefined negative sampling distribution and often is implemented as $p$, which is proportional to the item popularity. It has been shown that sampled softmax performs better than NCE ~\cite{nce} and negative sampling ones ~\cite{bpr,sampling_strategy} when the item number is large~\cite{cl-rec}. 

\noindent{\bf Contrastive Learning Loss (InfoNCE):} 
The key idea of contrastive learning is to contrast semantically similar (positive) and dissimilar (negative) pairs of data points, thus pulling the similar points closer and pushing the dissimilar points apart. There are several contrastive learning losses have been proposed~\cite{SimCLR,infonce,unbiased}, and among them, InfoNCE loss is one of the most well-known and has been adopted in the recommendation as follows: 

\begin{equation}\label{eq:infonce_0}
   \mathcal{L}_{info}= - \E_u \E_{i \sim p_u^+} \log \frac{exp(\hat{y}_{ui})}{exp(\hat{y}_{ui}) + \sum_{j=1; j \sim p_u^{-}}^N  exp(\hat{y}_{uj})}
\end{equation}

where $p_u^+$ ($p_u^-$) is the positive (negative) item distribution for user $u$. 
Note that unlike the sampled softmax, the correction from the sampling distribution is not present in the InfoNCE loss. In fact, this loss does not actually approximate the softmax probability measure $p(i|u)$ while it corresponds to the probability that {\em item $i$ is the positive item, and the remaining points are noise}, and $exp(\hat{y}_{ui})$ measures the density ration ($exp(\hat{y}_{ui}) \varpropto p(i|u)/p(i)$) ~\cite{infonce}. 
Again, the overall InfoNCE loss is the cross-entropy of identifying the positive items correctly, and it also shows maximize the mutual information between user $u$ and item $i$ ($I(i,u)$) and minimize $I(j,u)$ for item $j$ and $u$ being unrelated \citep{infonce}. 

In practice, as the true negative labels (so does all true positive labels) are not available,  negative items $j$ are typically drawn uniformly (or proportional to their popularity) from the entire item set $\mathcal{I}$ or $\mathcal{I}\backslash \MI^+_u$, i.e, $j \sim p_u$: 

\begin{equation}
   \widetilde{\mathcal{L}}_{info}= - \E_u \E_{i \sim p_u^+} \log \frac{exp(\hat{y}_{ui})}{exp(\hat{y}_{ui}) + \sum_{j=1; j \sim p_u}^N  exp(\hat{y}_{uj})}
\end{equation}

Note that this loss $\widetilde{\mathcal{L}}_{info}$ in practice  is sometimes treated as a softmax estimation, despite is not a proper estimator for the softmax loss~\cite{infonce,youtube}. In addition, the $\widetilde{\mathcal{L}}_{info}$ approximation easily leads to {\em sampling bias}: when sampling a large number of supposed negative items, they would include some positive items i.e., those top $K$ items in recommendation setting. It has been known such sampling bias can empirically lead to a significant performance drop for machine learning tasks~\cite{bias-debias,sampling_strategy}.

\vspace*{-2.0ex}
\subsection{Leveraging Debiased Contrastive InfoNCE}\label{subsec:debiased_infonce}
Can we reduce the gap between the approximation $\widetilde{\mathcal{L}}_{info}$ and ideal InfoNCE loss $\mathcal{L}_{info}$ and mitigate sampling bias?
In ~\cite{debiased}, a novel debiased estimator is proposed while only accessing positive items and unlabeled training data. Below, we demonstrate how this approach can be adopted in the recommendation setting. 

Given user $u$, let us consider the implicit (binary) preference probability (class distribution) of $u$: $\rho_u(c)$, where $\rho_u(c=+)=\tau^+_u$ represents the probability of user $u$ likes an item, and $\rho_u(c=-)=\tau^-_u=1-\tau^+_u$ corresponds to being not interested. In  the recommendation setting, we may consider $\rho_u(+)$ as the fraction of positive items, i.e., those top $K$ items, and  $\rho_u(-)$ are the fraction of true negative items.  Let the item class joint distribution be $p_u (i,c)=p_u(i|c)\rho_u(c)$. 
Then $p^+_u(i)=p_u(i|c=+)$ is the probability of observing $i$ as a positive item for user $u$ and $p_u^-(j)=p_u(j|c=-)$ the probability of a negative example. Given this, the distribution of an individual item of user $u$ is:
\begin{equation}
    \begin{split}
        p_u(i) = \tau^+_u \cdot p^+_u(i) + \tau^-_u \cdot p_u^-(i)
    \end{split}
\end{equation}

For implicit recommendation, we can assume the known positive items $\mathcal{I}_u^+$ are being uniformly sampled according to $p^+_u$. For $\tau^+_u$, we can count all the existing positive items ($r_{ui}=1$), and the unknown top $K$ items as being positive, thus, $\tau^+_u=(|\MI_{u}^+|+K)/|\mathcal{I}|$. Alternatively, we can assume the number of total positive items for $u$ is proportional to the number of existing (known) positive items, i.e, $\tau^+_u=(1+\alpha) |\MI_{u}^+|/|\mathcal{I}|$, where $\alpha\geq 0$. As mentioned earlier, the typical assumption of $p_u$ (the probability distribution of selection an item is also uniform). 
Then, we can represent $p^-_{u}$ using the total and positive sampling as: $p^-_{u}(i)=1/\tau^-_u (p_u(i)-\tau^+_u \cdot p^+_u(i))$.

Given this, let us consider the following ideal (debiased) InfoNCE loss: 
\begin{small}
\begin{equation}
\begin{split}
    &\overline{\mathcal{L}}_{Info}=- \E_{u} \E_{i \sim p_u^+}\Bigg[\log\frac{\exp(\hat{y}_{ui})}{\exp(\hat{y}_{ui}) + \lambda \E_{j\sim p_u^-} \exp({\hat{y}_{uj}}) }
    \Bigg] = \\
    &- \E_{u} \E_{i\sim p_u^+}\Bigg[\log\frac{\exp(\hat{y}_{ui})}{\exp(\hat{y}_{ui}) + \lambda \frac{1}{\tau^-}\Big( 
    \E_{t\sim p_u}[\exp(\hat{y}_{ut})] - \tau^+ \E_{j\sim p^+_u}[\exp(\hat{y}_{uj})]\Big) }
    \Bigg]
\end{split}
\end{equation}
\end{small}
Note that when $\lambda=N$, the typical InfoNCE loss $\mathcal{L}_{Info}$ (\cref{eq:infonce_0}) converges into $\overline{\mathcal{L}}_{Info}$ when $N \rightarrow \infty$.
Now, we can use $N$ samples from $p_u$ (not negative samples), an $M$ positive samples from $p_u^+$ to empirically estimate $\overline{\mathcal{L}}_{Info}$ following ~\cite{debiased}: 
\begin{small}
\begin{equation}
\begin{split}
    &\mathcal{L}^{Debiased}_{Info}=-\E_{u} \E_{i \sim p_u^+; j \sim p_u; k \sim p_u^+} \Bigg[\log\frac{\exp(\hat{y}_{ui})}{\exp(\hat{y}_{ui}) + \lambda f(\{j\}_{1}^N, \{k\}_{1}^M)}
    \Bigg], where \\
    & f(\{ j\}_{1}^N, \{k\}_{1}^M)= 
    \max \Bigl\{ \frac{1}{\tau_u^-} \Bigl( \frac{1}{N} \sum_{j=1;j\sim p_u}^N exp(f_{uj})- \tau_u^+\frac{1}{M} \sum_{k=1;k\sim p_u^+}^M exp(f_{uk}) \Bigr), e^{1/t}  \Bigr\}
\end{split}
\end{equation}
\end{small}
Here $f$ is constrained to be greater than its theoretical minimum $e^{-1/t}$ to prevent the negative number inside $\log$.
We will experimentally study this debiased InfoNCE in \cref{sec:experiment}. 

\subsection{Connection with MINE and BPR}
In the following, we first introduce Mutual Information Neural Estimator (MINE) loss and then discuss some common lower bounds for  InfoNCE and MINE, BPR as well as their relationship. Later in the \cref{sec:experiment}, we evaluate MINE against existing losses for recommendation and show it  performs surprisingly well. To the best of our knowledge,  this work is the first to study and apply MINE \citep{MINE} to the recommendation setting. 




 The Mutual Information Neural Estimator (MINE) loss is introduced in ~\cite{MINE}, which can directly measure the mutual information between user and item: 
 \begin{equation}
    \widehat{I(u,i)}=sup_{(v_u;v_i)} \E_{p_{u,i}}[\hat{y}_{ui}] - \log \E_{p_u \otimes p_i}[e^{\hat{y}_{ui}}]
\end{equation}
Here, the positive $\hat{y}_{ui}$ items sampled $n$ times according to the joint user-item distribution $p_{u,i}$. The negative items are sampled according to the empirical marginal  item distribution of $p_i$, which are proportional to their popularity (or in practice, sampled uniformly).
A simple adaption of MINE to the recommendation setting can simply exclude the positive item $i$ from the denominator of InfoNCE~\cite{infonce}: 

\begin{equation}\label{eq:mine}
\begin{split}
    \mathcal{L}_{mine}& =- \E_u \E_{i\sim p_u^+} \Bigg[\hat{y}_{ui}- \log \E_{j \sim p_i} [\exp \hat{y}_{uj}]\Bigg] 
\end{split}
\end{equation}

Its empirical loss can be written as (the factor of $1/N$ will not affect the optimization): 

\begin{equation}\label{eq:mine}
\begin{split}
    \widetilde{\mathcal{L}}_{mine}& =- \E_u \E_{i\sim p_u^+} \Bigg[\hat{y}_{ui}- \log \sum_{j=1; j \sim p_i}^N [\exp \hat{y}_{uj}]\Bigg] 
\end{split}
\end{equation}

\noindent{\bf Lower Bounds:} 
Given this, we can easily observe (based on Jensen's inequality): 
\begin{equation} 
\begin{split}
    &\widetilde{\mathcal{L}}_{info} = - \E_u \E_{i \sim p_u^+} \log \frac{exp(\hat{y}_{ui})}{exp(\hat{y}_{ui}) + \sum_{j=1; j \sim p_u}^N  exp(\hat{y}_{uj})} \geq \\
    & \widetilde{\mathcal{L}}_{mine} =  
    \E_u \E_{i\sim p_u^+} \log (\sum_{j=1; j\sim p_i}^N exp(\hat{y}_{uj} - \hat{y}_{ui})) \approx \\  
    & \E_u \E_{i\sim p_u^+} \log N \E_{j\sim p_i} exp(\hat{y}_{uj} - \hat{y}_{ui})) \geq  \\
    & \E_u \E_{i\sim p_u^+} \E_{j\sim p_i} \bigl( \hat{y}_{uj} - \hat{y}_{ui}  \bigr) + \log N 
    \geq \E_u \E_{i\sim p_u^+} \E_{j\sim p_i} \bigl( \hat{y}_{uj} - \hat{y}_{ui}  \bigr)
\end{split}
\end{equation}
Alternative, by using LogSumExp lower-bound~\cite{unbiased}, we also observe:  
\begin{equation} 
\begin{split}
    & \E_u \E_{i\sim p_u^+} \max_{j=1; j\sim p_i} ^N \bigl( 0, \hat{y}_{uj} - \hat{y}_{ui}  \bigr)  + \log (N+1) \geq \\
     & \widetilde{\mathcal{L}}_{info} \geq 
     \widetilde{\mathcal{L}}_{mine} =  
    \E_u \E_{i\sim p_u^+} \log (\sum_{j=1; j\sim p_i}^N exp(\hat{y}_{uj} - \hat{y}_{ui})) \geq \\ 
    &\E_u \E_{i\sim p_u^+} \max_{j=1; j\sim p_i} ^N \bigl( \hat{y}_{uj} - \hat{y}_{ui}  \bigr)
\end{split}
\end{equation}
Note that the second lower bound is tighter than the first one. We also observe the looser bound is also shared by the well-known pairwise BPR loss, which can be written as:

\begin{equation}
\begin{split}
&\widetilde{\mathcal{L}}_{bpr} = \E_u \E_{i\sim p_u^+} \sum_{j=1;j\sim p_i}^N -\log \sigma (\hat{y}_{ui} - \hat{y}_{uj})  \\
& =  \E_u \E_{i\sim p_u^+} \sum_{j=1; j\sim p_i}^N \log (1 + exp(\hat{y}_{uj} - \hat{y}_{ui})) 
\end{split}
\end{equation}

Following above analysis, we observe: 
\begin{equation}\label{eq:bound1}
\begin{split}
    &\widetilde{\mathcal{L}}_{info} = \E_u \E_{i \sim p_u^+} \log (1 + \sum_{j=1; j \sim p_i}^N  exp(\hat{y}_{uj}-\hat{y}_{ui})) \approx \\
    & \E_u \E_{i \sim p_u^+} \log (1 + N \E_{j \sim p_i}  exp(\hat{y}_{uj}-\hat{y}_{ui}))
    \geq \\
    & \E_u \E_{i \sim p_u^+} \E_{j \sim p_i} \log (1 + N exp(\hat{y}_{uj}-\hat{y}_{ui})) 
     \geq \\ 
    & \E_u \E_{i \sim p_u^+} \E_{j \sim p_i} \log (1/N + exp(\hat{y}_{uj}-\hat{y}_{ui})) + \log N \geq \\ 
    & \E_u \E_{i \sim p_u^+} \E_{j \sim p_i} \bigl(\hat{y}_{uj}-\hat{y}_{ui}\bigr)) + \log N  \geq \\
    &  \E_u \E_{i \sim p_u^+} \E_{j \sim p_i} \bigl(\hat{y}_{uj}-\hat{y}_{ui}\bigr))
\end{split}
\end{equation}
Thus, the BPR loss shares the same lower bound $E_u E_{i \sim p_u^+} \E_{j \sim p_i} \bigl(\hat{y}_{uj}-\hat{y}_{ui}\bigr))$ as the InfoNCE and MINE, which aims to maximize the average difference between a positive item score ($\hat{y}_{ui}$) and a negative item score ($\hat{y}_{uj}$). 

However, the tighter lower bound of BPR diverges from the InfoNCE and MINE loss using the LogSumExp lower bound: 
\begin{equation}\label{eq:bound2}
\begin{split}
&\widetilde{\mathcal{L}}_{bpr} =  \E_u \E_{i\sim p_u^+} \sum_{j=1; j\sim p_i}^N \log (1 + exp(\hat{y}_{uj} - \hat{y}_{ui})) \geq \\
& \E_u \E_{i\sim p_u^+} \sum_{j=1; j\sim p_i}^N 
max(0, \hat{y}_{uj} - \hat{y}_{ui}) 
\varpropto \\ & \E_u \E_{i\sim p_u^+} \E_{j\sim p_i}
max(0, \hat{y}_{uj} - \hat{y}_{ui})
\end{split}
\end{equation}
Note that the bound in \cref{eq:bound2} only sightly improves over the one in \cref{eq:bound1} and cannot reflect the bounds shared by InfoNCE and MINE, which aims to minimize the largest gain of a negative item (random) over a positive item ($\max_{j=1; j\sim p_i} ^N \bigl( \hat{y}_{uj} - \hat{y}_{ui}  \bigr)$). We also experimentally validate using the first bound $\E_u \E_{i \sim p_u^+} \E_{j \sim p_i} \bigl(\hat{y}_{uj}-\hat{y}_{ui}\bigr)$ leads to poor performance, indicating this can be a potentially underlying factor the relatively inferior performance of BRP comparing with other losses, such as softmax, InfoNCE, and MINE (\cref{sec:experiment}). 

\noindent{\textbf{MINE+ loss for Recommendation:}}
Utimately, we consider the following refined MINE loss format for recommendation which as we will show in \cref{sec:experiment}, can provide an additional boost comparing with the Vanilla formula (\cref{eq:mine}): 

\begin{equation}\label{eq:mine+}
\begin{split}
    \widetilde{\mathcal{L}}_{mine+}& =- \E_u \E_{i\sim p_u^+} \Bigg[\hat{y}_{ui}/t- \lambda \log \sum_{j=1; j \sim p_i}^N [\exp  (\hat{y}_{uj}/t]\Bigg] 
\end{split}
\end{equation}
Here, $\hat{y}_{ui}$ uses the cosine similarity between user and item latent vectors, $t$ serves as the temperature  (similar to other contrastive learning loss), and finally, $\lambda$ helps balance the weight between positive scores and negative scores.

\section{Debiased Pointwise Loss}\label{sec:linear}
Besides the typical listwise (softmax, InfoNCE) and pairwise (BPR) losses, another type of recommendation loss is the pointwise loss. Most of the linear (non-deep) models are based on pointwise losses, including the well-known iALS~\cite{hu2008collaborative,ials_revisiting}, EASE~\cite{ease}, and latest CCL~\cite{simplex}. Here, these loss functions aim to pull the estimated score $\hat{y}_{ui}$ closer to its default score $r_{ui}$. In all the existing losses, when the interaction is known, $i \in \MI^+_u$, $r_{ui}=1$; otherwise, the unlabeled (interaction unknown) items are treated as $r_{ui}=0$. 

Following the analysis of debiased contrastive learning, clearly, some of the positive items are unlabeled, and making them closer to $0$ is not the ideal option. In fact, if the optimization indeed reaches the targeted score, we actually do not learn any useful recommendations. A natural question is whether we can improve  the pointwise losses following the debiased contrastive learning paradigm ~\cite{debiased}. In addition, we will investigate if the popular linear methods, such as iALS and EASE, are debiased or not. 

\subsection{Debiased MSE and CCL}

Let us denote the generic single pointwise loss function $l^+(\hat{y}_{ui})$ and $l^-(\hat{y}_{ui})$, which measure how close the individual positive (negative) item score to their ideal target value. 

\noindent{\bf MSE single pointwise loss:}
Mean-Squared-Error (MSE) is one of the most widely used recommendation loss functions.  Indeed, all the earlier collaborative filtering (Matrix factorization) models and pointwise losses were  based on MSE (with different regularization). Some recent linear models, such as SLIM~\cite{slim} and EASE~\cite{ease}, are also based on MSE. 
Its single pointwise loss function is denoted as: 
\begin{equation}
\begin{cases}
    & l^+_{mse}(\hat{y}_{ui})=(1-\hat{y}_{ui})^2 \\
    & l^-_{mse}(\hat{y}_{ui})=(\hat{y}_{ui})^2
\end{cases}
\end{equation}

\noindent{\bf CCL single pointwise loss:}
The cosine contrastive loss (CCL) is recently proposed loss and has shown to be very effective for recommendation~\cite{}.
The original Contrastive Cosine Loss (CCL) \citep{simplex} can be written as:
\begin{equation}
    \begin{split}
        \mathcal{L}_{CCL}=\E_{u}
        \Bigg(\sum\limits_{i\in \mathcal{I}^+_u} (1-\hat{y}_{ui}) + \frac{w}{N}\sum\limits_{j\in\mathcal{I}}^N ReLU(\hat{y}_{uj}-\epsilon)\Bigg)
    \end{split}
\end{equation}
where $\hat{y}_{uj}$ is cosine similarity, $ReLU(x) = max(0, x)$ is the activation function, $w$ is the negative weight, $N$ is the number of negative samples and $\epsilon$ is margin.
Its single pointwise loss thus can be written as: 
\begin{equation}
\begin{cases}
    & l^+_{ccl}(\hat{y}_{ui})=1-\hat{y}_{ui} \\
    & l^-_{ccl}(\hat{y}_{ui})=ReLU(\hat{y}_{ui}-\epsilon)
\end{cases}
\end{equation}

Given this, we introduce the ideal pointwise loss for recommendation models (following the basic idea of (debiased) contrastive learning) : 

\bdefin {\bf (Ideal Pointwise Loss)}
The ideal pointwise loss function for recommendation models is the expected single pointwise loss for all positive items (sampled from $\rho_u(+)=
\tau^+$) and for negative items (sampled from $\rho_u(-)=
\tau^-$):  
\begin{small}
\begin{equation}
    \begin{split}
    \overline{\mathcal{L}}_{point} = \E_{u} \Bigg(\tau_u^+\E_{i\sim p_u^+} l^+(\hat{y}_{ui}) +
    \lambda\tau_u^-\E_{j\sim p_u^-} l^-(\hat{y}_{uj}) \Bigg)
    \end{split}
\end{equation}
\end{small}
\edefin

Why this is ideal? When the optimization reaches the potential minimal, we have for all positive items $i$ and all the negative items $j$ reaching all the minimum of individual loss ($l^+$ and $l^-$).  
Here, $\lambda$ help adjusts the balance between the positive and negative losses, similar to iALS ~\cite{hu2008collaborative,ials_revisiting} and CCL~\cite{simplex}. 
However, since $p_u^+$ and $p_u^-$ is unknown, we utilize the same debiasing approach as InfoNCE ~\cite{infonce}, also in \cref{subsec:debiased_infonce}, for the debiased MSE: 

\bdefin {\bf (Debiased Ideal pointwise loss)}
\begin{equation}
    \begin{split}
    &\mathcal{L}^{Debiased}_{point}= \E_{u} \Bigl( \tau_u^+\E_{i\sim p_u^+} l^+(\hat{y}_{ui}) + \tau_u^-\lambda \cdot \E_{j\sim p_u^-} l^-(\hat{y}_{ui}) \Bigr)\\
    & = \E_{u} \Big(\tau_u^+\E_{i\sim p_u^+}l^+(\hat{y}_{ui}) + \lambda\Big( 
    \E_{j\sim p_u}l^-(\hat{y}_{ui})  - \tau_u^+ \E_{k\sim k^+_u}l^-(\hat{y}_{uj}) \Big)\Big)
    \end{split}
\end{equation}
\edefin

In practice, assuming for each positive item, we sample $N$ positive items and $M$ negative items, then the empirical loss can be written as: 

\begin{equation}
    \begin{split}
    &\widetilde{\mathcal{L}}^{Debiased}_{point}= \E_{u} E_{i \sim p_u^+} \Bigl( \tau_u^+l^+(\hat{y}_{ui}) + \\ 
    &\lambda \Bigl( \frac{1}{N} \sum_{j=1;j\sim p_u}^N l^-(\hat{y}_{uj})- \tau_u^+\frac{1}{M} \sum_{k=1;k\sim p_u^+}^M l^-(\hat{y}_{uk}) \Bigr) \Bigr)
    \end{split}
\end{equation} 

As an example, we have the ideal $\overline{\mathcal{L}}_{CCL}$, debiased $\mathcal{L}^{debiased}_{CCL}$ , empirical debiased CCL $\widetilde{\mathcal{L}}^{Debiased}_{ccl}$ losses as follows: 

\begin{equation*}
    \begin{cases}
    & \overline{\mathcal{L}}_{CCL} = \E_{u} \Bigg(\tau_u^+\E_{i\sim p_u^+}(1-\hat{y}_{ui}) +
    \lambda\tau_u^-\E_{j\sim p_u^-}[ReLU(\hat{y}_{uj}-\epsilon)]\Bigg) \\
    &\mathcal{L}^{debiased}_{CCL} = \E_{u} \Bigg(\tau_u^+\E_{i\sim p_u^+}(1-\hat{y}_{ui}) \\  
    &+  \lambda\Big( 
    \E_{j\sim p_u}[ReLU(\hat{y}_{uj}-\epsilon)] - \tau_u^+ \E_{k\sim p^+_u}[ReLU(\hat{y}_{uk}-\epsilon)]\Big)\Bigg) \\
    &\widetilde{\mathcal{L}}^{Debiased}_{ccl}= \E_{u} E_{i \sim p_u^+} \Bigl( \tau_u^+(1-\hat{y}_{ui}) + \\ 
    &\lambda \Bigl( \frac{1}{N} \sum_{j=1;j\sim p_u}^N ReLU(\hat{f}_{uj}-\epsilon)- \tau_u^+\frac{1}{M} \sum_{k=1;k\sim p_u^+}^M ReLU(\hat{f}_{uj}-\epsilon) \Bigr) \Bigr)
    \end{cases}
\end{equation*}
We will study how (empirical) debiased MSE and CCL in the experimental result section. 

\subsection{iALS, EASE and their Debiasness}  
Here, we investigate how the debiased MSE loss will impact the solution of two (arguably) most popular linear recommendation models, iALS ~\cite{hu2008collaborative,ials_revisiting} and EASE~\cite{ease}. 
Rather surprisingly, we found the solvers of both models can absorb the debiased loss under their existing framework with reasonable conditions (details below). In other words, both iALS and EASE can be considered to be inherently debiased. 

To obtain the aforementioned insights, we first transform the debiased loss into the summation form being used in iALS and EASE. 
\begin{small}
\begin{equation*}
\begin{split}
&\mathcal{L}^{Debiased}_{mse} \approx 
\E_{u} \Bigg[ \frac{\tau^+_u} {|\mathcal{I}^+_u|}\sum\limits_{i\in \mathcal{I}_u^+}(\hat{y}_{ui}-1)^2 + \lambda\Bigg(\frac{1}{|\mathcal{I}|} \sum\limits_{t\in\mathcal{I}} \hat{y}_{ut}^2 -\frac{\tau^+_u}{|\mathcal{I}_u^+|}\sum\limits_{q\in \mathcal{I}_u^+}\hat{y}^2_{uq} \Bigg)\Bigg]\\
&=\sum\limits_{u} \Bigg[ \frac{1}{|\mathcal{I}|} c_u \sum\limits_{i\in \mathcal{I}_u^+}(\hat{y}_{ui}-1)^2 \  \ \ \ \ \ \ \ \ \ \ \ \ \ \text{where, }  c_u=\frac{|\mathcal{I}|}{|\mathcal{I}_u^+|} \tau_u^+
\\
&+ \lambda\Bigg(\frac{1}{|\mathcal{I}|} \sum\limits_{t\in\mathcal{I}} \hat{y}_{ut}^2 -\frac{1}{|\mathcal{I}|} c_u \sum\limits_{q\in \mathcal{I}_u^+}\hat{y}^2_{uq} \Bigg)\Bigg]\\
&\propto \sum\limits_{u} \Bigg[ c_u \sum\limits_{i\in \mathcal{I}_u^+}(\hat{y}_{ui}-1)^2 + \lambda\Bigg(\sum\limits_{t\in\mathcal{I}} \hat{y}_{ut}^2 - c_u\sum\limits_{q\in \mathcal{I}_u^+}\hat{y}^2_{uq} \Bigg)\Bigg]\\
&=\sum\limits_{u} \Bigg[ \sum\limits_{i\in \mathcal{I}_u^+}[c_u(\hat{y}_{ui}-1)^2 -c_u\lambda \hat{y}^2_{uq}]+ \lambda\sum\limits_{t\in\mathcal{I}} \hat{y}_{ut}^2 \Bigg]
\end{split}
\end{equation*}
\end{small}

\noindent{\bf Debiased iALS:}
Following the original iALS paper \citep{ials@hu2008}, and the latest revised iALS \citep{ials_revisiting,ials++}, the objective of iALS is given by \citep{ials_revisiting}:

\begin{equation}
    \begin{split}
    \mathcal{L}_{iALS}&=\sum\limits_{(u,i)\in S} (\hat{y}(u,i)-1)^2 + \alpha_0\sum\limits_{u\in U}\sum\limits_{i\in I}\hat{y}(u,i)^2\\
    &+\lambda\Bigg( \sum\limits_{u\in U}(|\mathcal{I}_u^+|+\alpha_0|\mathcal{I}|)^{\nu}||\mathbf{w}_u||^2 + \sum\limits_{i\in I}(|U_i^+|+\alpha_0|\mathcal{U}|)^{\nu}||\mathbf{h}_i||^2\Bigg)
    \end{split}
\end{equation}
where $\lambda$ is global regularization weight (hyperparameter) and $\alpha_0$ is unobserved weight (hyperparameter). $\mu$ is generally set to be $1$. 
Also, $\mathbf{w}_u$ and $\mathbf{h}_i$ are user and item vectors for user $u$ and item $i$, respectively. 

Now, we consider applying the debiased MSE loss to replace the original MSE loss (the first line in $\mathcal{L}_{iALS}$) with the second line unchanged: 

\begin{small}
\begin{equation*}
\begin{split}
&\mathcal{L}^{Debiased}_{iALS}
=\sum\limits_{u\in U} \Bigg[ \sum\limits_{i\in \mathcal{I}_u^+}[c_u(\hat{y}_{ui}-1)^2 -c_u\cdot \alpha_0 \cdot \hat{y}^2_{ui}]+ \alpha_0\sum\limits_{t\in I} \hat{y}_{ut}^2 \Bigg]\\
 &+\lambda\Bigg( \sum\limits_{u\in U}(|\mathcal{I}_u^+|+\alpha_0|\mathcal{I}|)^{\nu}||\mathbf{w}_u||^2 + \sum\limits_{i\in I}(|\mathcal{U}_i^+|+\alpha_0|\mathcal{U}|)^{\nu}||\mathbf{h}_i||^2\Bigg)
\end{split}
\end{equation*}
\end{small}

Then we have the following conclusion:
\bthm\label{th1}
For any debiased iALS loss $\mathcal{L}^{Debiased}_{iALS}$ with parameters $\alpha_0$ and $\lambda$ with constant $c_u$ for all users, there are original iALS loss with parameters  $\alpha_0^\prime$ and $\lambda^\prime$, which have the same closed form solutions (up to a constant factor) for fixing item vectors and user vectors, respectively. (proof in \cref{app:proof})
\ethm 

\noindent{\bf Debiased EASE:}
EASE~\cite{ease} has shown to be a simple yet effective recommendation model and evaluation bases (for a small number of items), and it aims to minimize:
\begin{equation*}
    \begin{split}
        &\mathcal{L}_{ease}=||X-XW||^2_F + \lambda||W||^2_F\\
        & s.t. \quad diag(W)=0
    \end{split}
\end{equation*}
It has a closed-form solution~\cite{ease}:
\begin{equation}
\begin{split}
&P = (X^TX+\lambda I)^{-1}\\
    &W^* = I-P\cdot dMat(diag(1\oslash P))
\end{split}
\end{equation}
where $\oslash$ denotes the elementwise division, and $diag$ vectorize the diagonal and $dMat$ transforms a vector to a diagonal matrix. 

To apply the debiased MSE loss into the EASE objective, let us first further transform $\mathcal{L}^{Debiased}_{mse}$ into parts of known positive items and parts of unknown items: 
\begin{equation*}
    \begin{split}
&\mathcal{L}^{Debiased}_{mse}=\sum\limits_{u} \Bigg[ \sum\limits_{i\in \mathcal{I}_u^+}[c_u(\hat{y}_{ui}-1)^2 -c_u\lambda \hat{y}^2_{uq}]+ \lambda\sum\limits_{t\in\mathcal{I}} \hat{y}_{ut}^2 \Bigg]\\
        &=\sum\limits_{u} \Bigg[ \sum\limits_{i\in \mathcal{I}_u^+}[c_u(\hat{y}_{ui}-1)^2 -c_u\lambda \hat{y}^2_{ui}]+ \lambda\sum\limits_{t\in\mathcal{I}_u^+} \hat{y}_{ut}^2 + \lambda\sum\limits_{p\in\mathcal{I}\backslash \mathcal{I}_u^+} \hat{y}_{ut}^2 \Bigg]\\
&=\sum\limits_{u} \Bigg[\sum\limits_{i\in \mathcal{I}_u^+}[c_u(\hat{y}_{ui}-1)^2+c_u \sum\limits_{p\in\mathcal{I}\backslash \mathcal{I}_u^+} \hat{y}_{ut}^2 \Bigg]\\
&+ \lambda (1-c_u)\sum\limits_{u}\sum\limits_{i\in \mathcal{I}_u^+}\hat{y}^2_{ui} + (\lambda-c_u) \sum\limits_{p\in\mathcal{I}\backslash \mathcal{I}_u^+} \hat{y}_{ut}^2 \\
 &=||\sqrt{C_u}(X-XW)||^2_F \\
        &-\lambda||X\odot \sqrt{C_u-I}XW||^2_F -||(1-X)\odot \sqrt{C_u-\lambda I}XW||^2_F
    \end{split}
\end{equation*}
Note that 
\begin{equation}
\sum\limits_{u}\sum\limits_{i\in \mathcal{I}}\hat{y}^2_{ui}=||XW||^2_F
\end{equation}

To find the closed-form solution, we further restrict $\lambda=1$ and consider $c_u$ as a constant (always $>1$), thus, we have the following simplified format: 


\begin{equation}
    \begin{split}
    \mathcal{L}^{Debiased}_{mse}=&||\sqrt{C_u}(X-XW)||^2_F
        -||\sqrt{C_u-I}XW||^2_F\\
        &=||X-XW||^2_F
        -\alpha ||XW||^2_F 
    \end{split}
\end{equation}
where, $\alpha=c_u-1$. 
Note that if we only minimize this objective function $\mathcal{L}^{Debiased}_{mse}$, we have the closed form: 
$$W = ((1-\lambda )X^TX)^{-1}X^TX.$$ 

Now, considering the debiased version of EASE:  
\begin{equation*}
    \begin{split}
        &\mathcal{L}^{Debiased}_{ease}=||X-XW||^2_F-\alpha||XW||^2_F + \lambda||W||^2_F\\
        & s.t. \quad diag(W)=0
    \end{split}
\end{equation*}
It has the following closed-form solution (by similar inference as EASE~\cite{ease}):


\begin{equation}
\begin{split}
\widehat{W}&=\frac{1}{1-\alpha} (I -\hat{P}\cdot dMat( \vec{1}\oslash diag(\hat{P}) ), \\ 
where \ \  & \hat{P}= (X^TX+\frac{\lambda}{1-\alpha} I)^{-1}
\end{split}
\end{equation}
Now, we can make the following observation: 
\bthm
For any debiased EASE loss $\mathcal{L}^{Debiased}_{ease}$ with parameters $\alpha$ and $\lambda$ with constant $c_u>1$ for all users, there are original EASE loss with parameter,  $\lambda^\prime$, which have the same closed form solutions EASE (up to constant factor). 
\ethm 
\bproof
Following above analysis and let 
$\lambda^\prime=\frac{\lambda}{(1-\alpha)c_u}$.
\eproof 

These results also indicate the sampling based approach to optimize the debiased $MSE$ loss may also be rather limited. In the experimental results, we will further validate this conjecture on debiased MSE loss. 

\section{Experiment}
\label{sec:experiment}

\begin{table}
\caption{Statistics of the datasets.}
\label{tab:dataset}
\begin{tabular}{c|c|c|c}
\hline { Dataset } & User \# & Item \# & Interaction \#  \\
\hline 
\hline Yelp2018 & 31,668 & 38,048 & $1,561,406$  \\
\hline Gowalla & 29,858 & 40,981 & $1,027,370$  \\
\hline Amazon-Books & 52,643 & 91,599 & $2,984,108$  \\
\hline
\end{tabular}
\end{table}

\begin{table*}[]
\caption{The performance of different loss functions w.r.t its debiased one on MF (Matrix Factorization). The unit of the metric values is $\%$. We also present and highlight the relative improvement (RI) of the debiased loss function with its vanilla (biased) counterpart.}
\label{tab:debias}
\begin{tabular}{|cc|cc|cc|cc|c|}
\hline
\multicolumn{2}{|c|}{\multirow{2}{*}{Loss}}                                                                                                                           & \multicolumn{2}{c|}{Yelp}                              & \multicolumn{2}{c|}{Gowalla}                             & \multicolumn{2}{c|}{Amazon-Books}                        & \multirow{2}{*}{Average RI\%} \\ \cline{3-8}
\multicolumn{2}{|c|}{}                                                                                                                                                & \multicolumn{1}{c|}{Recall@20}       & NDCG@20         & \multicolumn{1}{c|}{Recall@20}        & NDCG@20          & \multicolumn{1}{c|}{Recall@20}        & NDCG@20          &                               \\ \hline
\multicolumn{1}{|c|}{\multirow{3}{*}{CCL}}                                                   & biased                                                                 & \multicolumn{1}{c|}{6.91}            & 5.67            & \multicolumn{1}{c|}{18.17}            & 14.61            & \multicolumn{1}{c|}{5.27}             & 4.22             & -                             \\ \cline{2-9} 
\multicolumn{1}{|c|}{}                                                                       & debiased                                                               & \multicolumn{1}{c|}{6.98}            & 5.71            & \multicolumn{1}{c|}{18.42}            & 14.97            & \multicolumn{1}{c|}{5.49}             & 4.40             & -                             \\ \cline{2-9} 
\multicolumn{1}{|c|}{}                                                                       & RI\%                                                                   & \multicolumn{1}{c|}{\textbf{1.01\%}} & \textbf{0.71\%} & \multicolumn{1}{c|}{\textbf{1.38\%}}  & \textbf{2.46\%}  & \multicolumn{1}{c|}{\textbf{4.17\%}}  & \textbf{4.27\%}  & \textbf{2.33\%}               \\ \hline
\multicolumn{1}{|c|}{\multirow{3}{*}{MSE}}                                                   & biased                                                                 & \multicolumn{1}{c|}{5.96}            & 4.95            & \multicolumn{1}{c|}{14.27}            & 12.39            & \multicolumn{1}{c|}{3.36}             & 2.68             & -                             \\ \cline{2-9} 
\multicolumn{1}{|c|}{}                                                                       & debiased                                                               & \multicolumn{1}{c|}{6.05}            & 5.00            & \multicolumn{1}{c|}{14.98}            & 12.64            & \multicolumn{1}{c|}{3.35}             & 2.61             & -                             \\ \cline{2-9} 
\multicolumn{1}{|c|}{}                                                                       & RI\%                                                                   & \multicolumn{1}{c|}{\textbf{1.51\%}} & \textbf{1.01\%} & \multicolumn{1}{c|}{\textbf{4.98\%}}  & \textbf{2.02\%}  & \multicolumn{1}{c|}{\textbf{-0.30\%}} & \textbf{-2.61\%} & \textbf{1.10\%}               \\ \hline
\multicolumn{1}{|c|}{\multirow{3}{*}{InfoNCE}}                                               & biased                                                                 & \multicolumn{1}{c|}{6.54}            & 5.36            & \multicolumn{1}{c|}{16.45}            & 13.43            & \multicolumn{1}{c|}{4.60}             & 3.56             & -                             \\ \cline{2-9} 
\multicolumn{1}{|c|}{}                                                                       & debiased                                                               & \multicolumn{1}{c|}{6.66}            & 5.45            & \multicolumn{1}{c|}{16.57}            & 13.72            & \multicolumn{1}{c|}{4.65}             & 3.66             & -                             \\ \cline{2-9} 
\multicolumn{1}{|c|}{}                                                                       & RI\%                                                                   & \multicolumn{1}{c|}{\textbf{1.83\%}} & \textbf{1.68\%} & \multicolumn{1}{c|}{\textbf{0.73\%}}  & \textbf{2.16\%}  & \multicolumn{1}{c|}{\textbf{1.09\%}}  & \textbf{2.81\%}  & \textbf{1.72\%}               \\ \hline
\multicolumn{1}{|c|}{\multirow{4}{*}{\begin{tabular}[c]{@{}c@{}}MINE\\ (ours)\end{tabular}}} & MINE                                                                   & \multicolumn{1}{c|}{6.56}            & 5.37            & \multicolumn{1}{c|}{16.93}            & 14.28            & \multicolumn{1}{c|}{5.00}             & 3.93             & -                             \\ \cline{2-9} 
\multicolumn{1}{|c|}{}                                                                       & \begin{tabular}[c]{@{}c@{}}RI\% \\ (w.r.t baised infoNCE)\end{tabular} & \multicolumn{1}{c|}{\textbf{0.31\%}} & \textbf{0.19\%} & \multicolumn{1}{c|}{\textbf{2.92\%}}  & \textbf{6.33\%}  & \multicolumn{1}{c|}{\textbf{8.70\%}}  & \textbf{10.39\%} & \textbf{4.80\%}               \\ \cline{2-9} 
\multicolumn{1}{|c|}{}                                                                       & MINE+                                                                  & \multicolumn{1}{c|}{7.12}            & 5.86            & \multicolumn{1}{c|}{18.13}            & 15.31            & \multicolumn{1}{c|}{5.18}             & 4.05             & -                             \\ \cline{2-9} 
\multicolumn{1}{|c|}{}                                                                       & \begin{tabular}[c]{@{}c@{}}RI\% \\ (w.r.t baised infoNCE)\end{tabular} & \multicolumn{1}{c|}{\textbf{8.87\%}} & \textbf{9.33\%} & \multicolumn{1}{c|}{\textbf{10.21\%}} & \textbf{14.00\%} & \multicolumn{1}{c|}{\textbf{12.61\%}} & \textbf{13.76\%} & \textbf{11.46\%}              \\ \hline
\end{tabular}
\end{table*}

In this section, we experimentally study the most widely used loss functions in the existing recommendation models and their corresponding debiased loss as well as newly proposed MINE loss under recommendation settings. This could help better understand and compare the resemblance and discrepancy between different loss functions, and understand their debiasing benefits.  Specifically, we would like to answer the following questions:
\begin{itemize}
\item Q1. How does the bias introduced by the existing negative sampling impact model performance and how do our proposed debiased losses perform compared to traditional biased ones?
\item Q2. How does our proposed MINE objective function in Section \ref{sec:softmax} compare to widely used ones?
\item Q3. What is the effect of different hyperparameters on the various debiased loss functions?
\end{itemize}

\subsection{Experimental Setup}

\subsubsection{Datasets}
We use three datasets, \textbf{Amazon-Books}, \textbf{Yelp2018} and \textbf{Gowalla} commonly used by a number of recent studies \citep{simplex,lightgcn,ngcf,sgl-ed}. We obtain the publicly available processed data and follow the same setting as \citep{ngcf,lightgcn,simplex}.

\subsubsection{Evaluation Metrics} Consistent with benchmarks \citep{ngcf,lightgcn,simplex}, we evaluate the performance by $Recall@20$ and $NDCG@20$ over all items \citep{walid@sample}.

\subsubsection{Models}
This paper focuses on the study of loss functions, which is architecture agnostic. Thus, we evaluate our loss functions on top of the simplest one - Matrix Factorization (MF) in this study. By eliminating the influence of the architecture, we can compare the power of various loss functions fairly. 
Meantime, we compare the performance of the proposed methods with a few classical machine learning models as well as advanced deep learning models, including, iALS \citep{ials_revisiting}, MF-BPR\citep{bpr}, MF-CCL \citep{simplex}, YouTubeNet \citep{youtube}, NeuMF \citep{ncf}, LightGCN \citep{lightgcn}, NGCF \citep{ngcf}, CML \citep{cml}, PinSage \citep{pinsage}, GAT \citep{gat}, MultiVAE \citep{multi-vae}, SGL-ED \citep{sgl-ed}.
Those comparison can help establish that the loss functions with basic MF can produce strong and robust baselines for recommendation community to evaluate more advanced and sophisticated models. 

\subsubsection{Loss functions}
As the core of this paper, we focus on some most widely used loss functions and their debiased counterparts, including InfoNCE \citep{infonce}, MSE \citep{ials@hu2008,ials_revisiting}, and state-of-the-art CCL \citep{simplex}, as well as new proposed MINE \citep{MINE}.

\subsubsection{Reproducibility}

To implement reliable MF-based baselines (with CCL, MSE, InfoNCE objectives), we search a few hyperparameters and present the best one. By default, we set the batch size of training as $512$. Adam optimizer learning rate is initially set as $1e-4$ and reduced by $0.5$ on the plateau, early stopped until it arrives at $1e-6$. For the cases where negative samples get involved, we would set it as $800$ by default. We search the global regularization weight between $1e-9$ to $1e-5$ with an increased ratio of $10$. 

For the exact hyperparameter settings of the model, we would present it in \cref{app:reproduce} and our anonymous code is available at \url{https://anonymous.4open.science/r/KDD-2023-Anonymous-5F83/README.md}

\begin{table*}[]
\caption{Perfomance comparison to widely used models. We highlight the top-3 best models in each column. Results of models marked with $*$ are duplicated from \citep{simplex} for consistency. For a fair comparison of the MF-based models marked with $**$ are reproduced by ourselves with thorough parameter searching detailed in \cref{app:reproduce}.}

\label{tab:main}
\begin{tabular}{|ccccccc|}
\hline
\multicolumn{1}{|c|}{\multirow{2}{*}{Model}} & \multicolumn{2}{c|}{Yelp}                                               & \multicolumn{2}{c|}{Gowalla}                                              & \multicolumn{2}{c|}{Amazon-Books}                  \\ \cline{2-7} 
\multicolumn{1}{|c|}{}                       & \multicolumn{1}{c|}{Recall@20}     & \multicolumn{1}{c|}{NDCG@20}       & \multicolumn{1}{c|}{Recall@20}      & \multicolumn{1}{c|}{NDCG@20}        & \multicolumn{1}{c|}{Recall@20}     & NDCG@20       \\ \hline
\multicolumn{7}{|c|}{Deep Learning Based}                                                                                                                                                                                                               \\ \hline
\multicolumn{1}{|c|}{YouTubeNet* \citep{youtube}}            & \multicolumn{1}{c|}{6.86}          & \multicolumn{1}{c|}{\textbf{5.67}} & \multicolumn{1}{c|}{17.54}          & \multicolumn{1}{c|}{14.73}          & \multicolumn{1}{c|}{5.02}          & 3.88          \\ \hline
\multicolumn{1}{|c|}{NeuMF* \citep{ncf}}                 & \multicolumn{1}{c|}{4.51}          & \multicolumn{1}{c|}{3.63}          & \multicolumn{1}{c|}{13.99}          & \multicolumn{1}{c|}{12.12}          & \multicolumn{1}{c|}{2.58}          & 2             \\ \hline
\multicolumn{1}{|c|}{CML* \citep{cml}}                   & \multicolumn{1}{c|}{6.22}          & \multicolumn{1}{c|}{5.36}          & \multicolumn{1}{c|}{16.7}           & \multicolumn{1}{c|}{12.92}          & \multicolumn{1}{c|}{\textbf{5.22}} & 4.28          \\ \hline
\multicolumn{1}{|c|}{MultiVAE* \citep{multi-vae}}              & \multicolumn{1}{c|}{5.84}          & \multicolumn{1}{c|}{4.5}           & \multicolumn{1}{c|}{16.41}          & \multicolumn{1}{c|}{13.35}          & \multicolumn{1}{c|}{4.07}          & 3.15          \\ \hline
\multicolumn{1}{|c|}{LightGCN* \citep{lightgcn}}              & \multicolumn{1}{c|}{6.49}          & \multicolumn{1}{c|}{5.3}           & \multicolumn{1}{c|}{\textbf{18.3}}  & \multicolumn{1}{c|}{\textbf{15.54}} & \multicolumn{1}{c|}{4.11}          & 3.15          \\ \hline
\multicolumn{1}{|c|}{NGCF* \citep{ngcf}}                  & \multicolumn{1}{c|}{5.79}          & \multicolumn{1}{c|}{4.77}          & \multicolumn{1}{c|}{15.7}           & \multicolumn{1}{c|}{13.27}          & \multicolumn{1}{c|}{3.44}          & 2.63          \\ \hline
\multicolumn{1}{|c|}{GAT* \citep{gat}}                   & \multicolumn{1}{c|}{5.43}          & \multicolumn{1}{c|}{4.32}          & \multicolumn{1}{c|}{14.01}          & \multicolumn{1}{c|}{12.36}          & \multicolumn{1}{c|}{3.26}          & 2.35          \\ \hline
\multicolumn{1}{|c|}{PinSage* \citep{pinsage}}               & \multicolumn{1}{c|}{4.71}          & \multicolumn{1}{c|}{3.93}          & \multicolumn{1}{c|}{13.8}           & \multicolumn{1}{c|}{11.96}          & \multicolumn{1}{c|}{2.82}          & 2.19          \\ \hline
\multicolumn{1}{|c|}{SGL-ED* \citep{sgl-ed}}                & \multicolumn{1}{c|}{6.75}          & \multicolumn{1}{c|}{5.55}          & \multicolumn{1}{c|}{-}              & \multicolumn{1}{c|}{-}              & \multicolumn{1}{c|}{4.78}          & 3.79          \\ \hline
\multicolumn{7}{|c|}{MF based}                                                                                   \\ \hline
\multicolumn{1}{|c|}{iALS** \citep{ials_revisiting}}                & \multicolumn{1}{c|}{6.06}          & \multicolumn{1}{c|}{5.02}          & \multicolumn{1}{c|}{13.88}          & \multicolumn{1}{c|}{12.24}          & \multicolumn{1}{c|}{2.79}          & 2.25        \\
\hline
\multicolumn{1}{|c|}{MF-BPR** \citep{bpr}}                & \multicolumn{1}{c|}{5.63}          & \multicolumn{1}{c|}{4.60}          & \multicolumn{1}{c|}{15.90}          & \multicolumn{1}{c|}{13.62}          & \multicolumn{1}{c|}{3.43}          & 2.65         \\ \hline
\multicolumn{1}{|c|}{MF-CCL** \citep{simplex}}               & \multicolumn{1}{c|}{\textbf{6.91}} & \multicolumn{1}{c|}{\textbf{5.67}} & \multicolumn{1}{c|}{\textbf{18.17}} & \multicolumn{1}{c|}{14.61}          & \multicolumn{1}{c|}{\textbf{5.27}} & \textbf{4.22} \\ \hline
\multicolumn{1}{|c|}{MF-CCL-debiased (ours)} & \multicolumn{1}{c|}{\textbf{6.98}} & \multicolumn{1}{c|}{\textbf{5.71}} & \multicolumn{1}{c|}{\textbf{18.42}} & \multicolumn{1}{c|}{\textbf{14.97}} & \multicolumn{1}{c|}{\textbf{5.49}} & \textbf{4.40} \\ \hline
\multicolumn{1}{|c|}{MF-MINE+ (ours)}        & \multicolumn{1}{c|}{\textbf{7.12}} & \multicolumn{1}{c|}{\textbf{5.86}} & \multicolumn{1}{c|}{18.13}          & \multicolumn{1}{c|}{\textbf{15.31}} & \multicolumn{1}{c|}{5.18}          & \textbf{4.06} \\ \hline
\end{tabular}
\end{table*}

\subsection{Q1. (Biased) Loss vs Debiased Loss}
Here, we seek to examine the negative impact of bias introduced by negative sampling and assess the effectiveness of our proposed debiased loss functions. We utilize Matrix Factorization (MF) as the backbone and compare the performance of the original (biased) version and the debiased version for the loss functions of $CCL$ \citep{simplex}, $MSE$ \citep{ials@hu2008,ials_revisiting}, and $InfoNCE$ \citep{infonce}.   Table \ref{tab:debias} shows the biased loss vs debiased loss (on top of the MF) results. 

The results show that the debiased methods consistently outperform the biased ones for the CCL and InfoNCE loss functions in all cases. In particular, the debiased method exhibits remarkable improvements of up to over $4\%$ for the CCL loss function on the $Amazon-Books$ dataset. This highlights the adverse effect of bias in recommendation quality and demonstrates the success of applying contrastive learning based debiasing methods in addressing this issue.
In addition, despite the robustness with respect to the biased MSE loss for iALS and EASE (as being proof in \cref{sec:linear}, the debiased MSE still provides slight gain over the existing MSE loss. Finally, we note that the performance gained by the debiasing losses is also consistent with the results over other machine learning tasks as shown in ~\cite{debiased}.

\subsection{Q2. Performance of MINE $\backslash$ MINE+}
Next, we delve into the examination of the newly proposed Mutual Information Neural Estimator (MINE) loss function. As we mentioned before, to our best knowledge, this is the first study utilizing and evaluating MINE loss in recommendation settings. We compare MINE with other loss functions, all set MF as the backbone, as displayed in Table \ref{tab:debias},. 
Our results show the surprising effectiveness of MINE-type losses.
In particular, the basic MINE is shown close to $5\%$ average lifts over the (biased) InfoNCE loss, a rather stand loss in recommendation models. MINE+ demonstrates comparable or superior performance compared to the state-of-the-art CCL loss function. Furthermore, MINE+ consistently achieves better results than InfoNCE with a maximum improvement of 14\% (with an average improvement over $11\%$. 

These findings highlight the potential of MINE as an effective objective function for recommendation models and systems.  
Noting that our MINE+ loss is model agnostic, which can be applied to any representation-based model.
But using it with the basic MF already provides strong and robust baselines for evaluating recommendation models (and systems). 

Furthermore, we compare MF-MINE+ as well as our debiased MF-CCL model with some advanced machine learning and deep learning models in \cref{tab:main}. We can find these two methods perform best or second best in most cases, affirming the superiority of the MINE+ loss and debiased CCL.


\subsection{Q3. Hyperparameter analysis}
\begin{figure}
    \centering
    \includegraphics[width=0.95\linewidth]{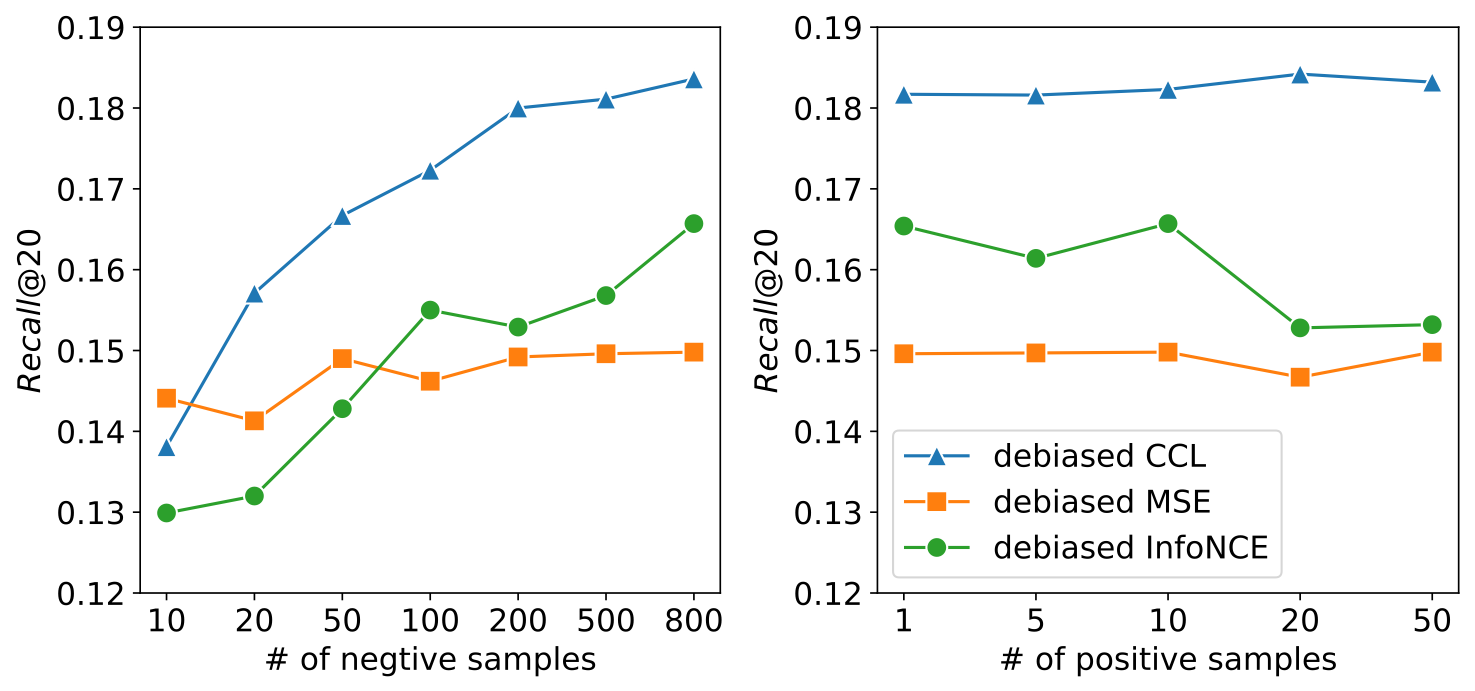}
    \vspace{-5pt}
    \caption{Effect of number of positive and negative samples on $Gowalla$}
    \label{fig:hyper_pos_neg}
\end{figure}

\begin{figure}
    \centering
    \includegraphics[width=0.95\linewidth]{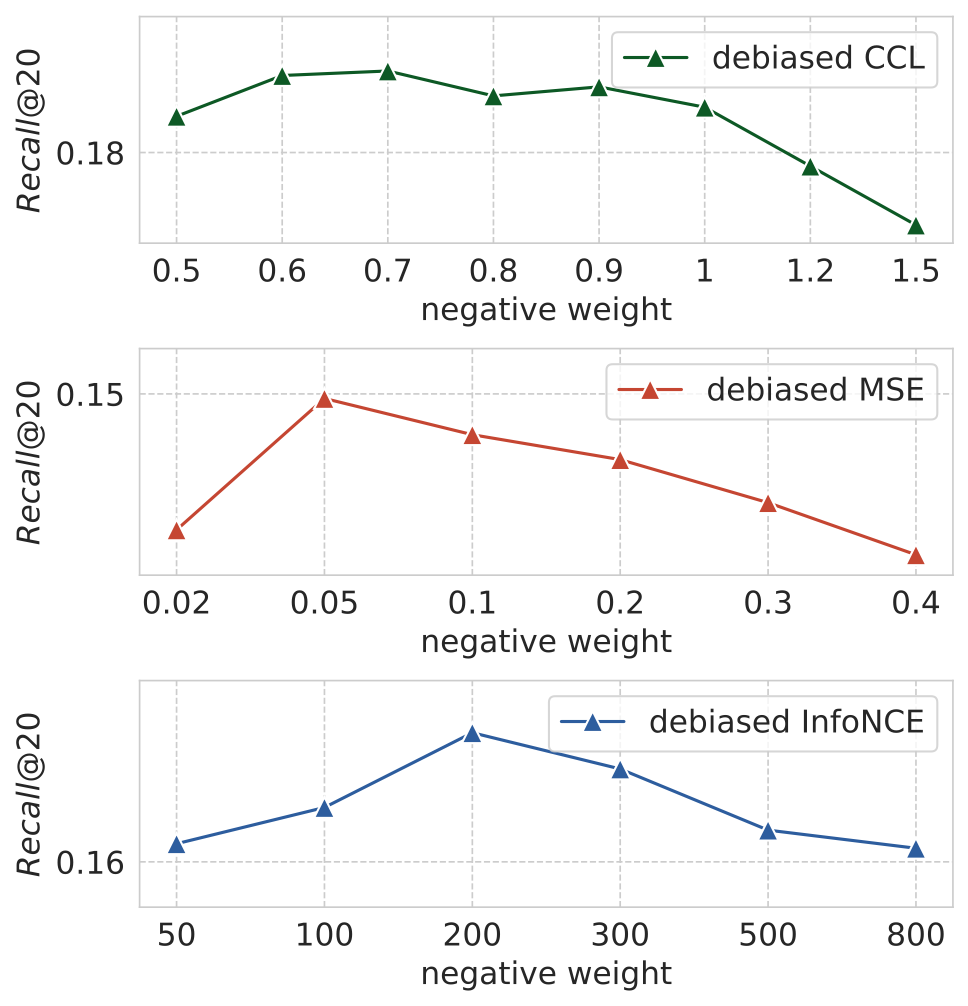}
    \vspace{-5pt}
    \caption{Effect of negative weight on $Gowalla$}
    \label{fig:hyper_lambda}
\end{figure}

Here, we analyze the impacts of three key hyperparameters in the debiased framework: negative weight $\lambda$, number of negative samples, and number of positive samples as shown in \cref{fig:hyper_pos_neg,fig:hyper_lambda}.

The left side of \cref{fig:hyper_pos_neg} reveals that, in general, increasing the number of negative samples leads to better performance in models. The MSE loss requires fewer negative samples to reach a stable plateau but performs worse when the number of negative samples increases. The right side of \cref{fig:hyper_pos_neg} shows that the number of positive samples in the debiased formula has a minor impact on performance compared to negative samples.

In \cref{fig:hyper_lambda}, we illustrate how the negative weight affects the performance of the debiased CCL, MSE, and InfoNCE losses. In general, we would expect the $\cap$ shaped curve for quality metric ($Recall@20$ here) for the parameters, which is consistent with our curves. The optimal negative weight for these loss functions locates at different orders of magnitude due to their intrinsic differences of loss functions. In general, the negative weights of MSE and CCLE are relatively small and the negative weights of InfoNCE are in the order of hundreds and related to the number of negative samples (\cref{app:reproduce}). 

\begin{figure}
    \centering
    \includegraphics[width=0.95\linewidth]{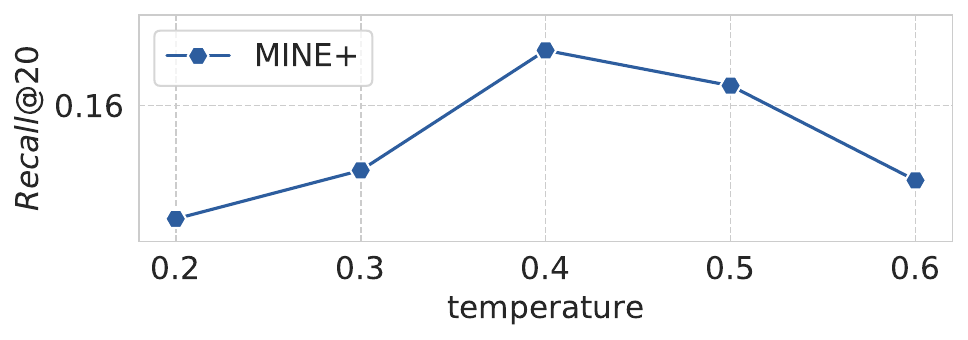}
    \caption{Effect of temperature on $Gowalla$}
    \label{fig:temp}
\end{figure}

In \cref{fig:temp}, we investigate how temperature can affect the performance of $MINE+$ loss function. In general, it would achieve optimal around $0.5$ for all datasets (see \cref{app:reproduce}).

\section{Related Work}
\label{sec:related}
\subsection{Objectives of Implicit Collaborative Filtering}

Implicit feedback has been popular for decades in Recommender System \citep{rendle2021item} since Hu et al. first proposed iALS \citep{ials@hu2008}, where a second-order pointwise objective - Mean Square Error (MSE) was adopted to optimize user and item embeddings alternatively. Due to its effectiveness and efficiency, there is a large number of works that take the MSE or its variants as their objectives, spanning from Matrix Factorization (MF) based models \citep{ials_revisiting,ials++,eals}, to regression-based models, including SLIM \citep{slim}, EASE \citep{ease,edlae}, etc. He et al. treat collaborative filtering  as a binary classification task and apply the pointwise objective - Binary Cross-Entropy Loss (BCE) onto it \citep{ncf}. CML utilized the pairwise hinge loss onto the collaborative filtering scenario \citep{cml}. MultiVAE \citep{multi-vae} utilizes multinomial likelihood estimation. Rendle et al. proposed a Bayesian perspective pairwise ranking loss - BPR, in the seminal work \citep{bpr}. YouTubeNet posed a recommendation as an extreme multiclass classification problem and apply the Softmax cross-entropy (Softmax) \citep{youtube,sample-softmax}. Recently, inspired by the largely used contrastive loss in the computer vision area, \citep{simplex} proposed a first-order based Cosine Contrastive Loss (CCL), where they maximize the cosine similarity between the positive pairs (between users and items) and minimize the similarity below some manually selected margin.

\subsection{Contrastive Learning}
Recently, Contrastive Learning (CL) has become a prominent optimizing framework in deep learning \citep{SimCLR,infonce}. The motivation behind CL is to learn representations by contrasting positive and negative pairs as well as maximize the positive pairs, including data augmentation \citep{SimCLR} and multi-view representations \citep{multi-cl}, etc. 
Chuang et al. proposed a new unsupervised contrastive representation learning framework, targeting to minimize the bias introduced by the selection of the negative samples. This debiased objective consistently improved its counterpart - the biased one in various benchmarks \citep{debiased}. Lately, by introducing the margin between positive and negative samples, Belghazi et al. present a supervised contrastive learning framework that is robust to biases \citep{unbiased}.
In addition, a mutual information estimator - MINE \citep{MINE}, closely related to the contrastive learning loss InfoNCE\citep{infonce}, has demonstrated its efficiency in various optimizing settings. \citep{alignment-uniformity} study the alignment and uniformity of user, item embeddings from the perspective of contrastive learning in the recommendation. CLRec \citep{cl-rec} design a new contrastive learning loss which is equivalent to using inverse propensity weighting to reduce exposure bias of a large-scale system. 



\subsection{Negative Sampling Strategies for Item Recommendation}
In most real-world scenarios, only positive feedback is available which brings demand for negative signals to avoid trivial solutions during training recommendation models. Apart from some non-sampling frameworks like iALS \citep{ials@hu2008}, Multi-VAE \citep{multi-vae}, the majority of models \citep{negative_sampling_review,lightgcn,ncf,bpr,rendle2021item,sampling_strategy} would choose to sample negative items for efficiency consideration. Uniform sampling is the most popular one \citep{bpr} which assumes uniform prior distribution. Importance sampling is another popular choice \citep{importance}, which chooses negative according to their frequencies. Adaptive sampling strategy keep tracking and picking the negative samples with higher scores \citep{aobpr,dynamic}. Another similar strategy is SRNS \citep{robustsample}, which tries to find and optimize false negative samples 
In addition, NCE \citep{nce} approach for negative sampling has also been adopted in the recommendation system\citep{sampling_strategy}. 


\subsection{Bias and Debias in Recommendation}
In the recommender system, there are various different bias \citep{bias-debias}, 
including item exposure bias, popularity bias, position bias, selection bias, etc.
These issues have to be carefully handled otherwise it would lead to unexpected results, such as inconsistency of online and offline performance, etc. Marlin et al. \citep{select-bias} validate the existence of selection bias by conducting a user survey, which indicates the discrepancy between the observed rating data and all data. Since users are generally exposed to a small set of items, unobserved data doesn't mean negative preference, this is so-called exposure bias \citep{expose-bias}. Position bias \citep{position-bias} is another commonly 
encountered one, where users are more likely to select items that display in a desirable place or higher position on a list.

Steck et al. \citep{steck-debias} propose an unbiased metric to deal with selection bias. iALS \citep{ials@hu2008} put some weights on unobserved data which helps improve the performance and deal with exposure bias. In addition, the sampling strategy would naturally diminish the exposure bias during model training \citep{bias-debias,bpr}. EXMF \citep{user-bias} is a user exposure simulation model that can also help reduce exposure bias.
To reduce position bias, some models \citep{dbcl,aecc} make assumptions about user behaviors and estimate true relevance instead of directly using clicked data. \citep{unbiased-pair} proposed an unbiased pairwise BPR estimator and further provide a practical technique to reduce variance. Another widely used method to correct bias is the utilization of propensity score \citep{unbiased-propensity,unbiased-rec-miss} where each item is equiped with some position weight.

\section{conclusion}
\label{sec:conclusion}
In this paper, we conduct a comprehensive analysis of recommendation loss functions from the perspective of contrastive learning. 
We introduce a list of debiased losses and  new mutual information-based loss functions - MINE and MINE+ to recommendation setting. We  also show how the point-wise loss functions can be debiased and certified both iALS and EASE are inherently debiased. 
The empirical experimental results demonstrate the debiased losses and new information losses outperform the existing (biased) ones. 
In the future, we would like to investigate how the loss functions work with more sophisticated neural architectures, and help discover more effective recommendation models by leveraging these losses. 



\section*{Acknowledgement}

This research was partially funded by the National Science Foundation through grants IIS-2142675, IIS-2142681, and III-2008557. Additional support came from a collaborative research agreement between Kent State University and iLambda Inc.

\bibliographystyle{ACM-Reference-Format}
\bibliography{sample-base}

\clearpage
\appendix
\section{reproducibility}\label{app:reproduce}


In \cref{tab:mine-repro}, we list the details of the hyperparameters for reproducing the results of $MINE+$ in \cref{tab:main,tab:debias}.

In \cref{tab:ccl-repro}, we list the details of the hyperparameters for reproducing the results of debiased $CCL$ in \cref{tab:debias}.

\begin{table}[]
\caption{Hyperparameter details of $MINE+$ loss}
\label{tab:mine-repro}
\begin{tabular}{|c|c|c|c|}
\hline
\multicolumn{1}{|l|}{} & Yelp2018 & Gowalla & Amazon-Books \\ \hline
negative weight        & 1.1      & 1.2     & 1.1          \\ \hline
temperature            & 0.5      & 0.4     & 0.4          \\ \hline
regularization         & 1e-8       & 1e-8      & 1e-8           \\ \hline
number of negative     & 800      & 800     & 800          \\ \hline
\end{tabular}
\end{table}

\begin{table}[]
\caption{Hyperparameter details of debiased $CCL$ loss}
\label{tab:ccl-repro}
\begin{tabular}{|c|c|c|c|}
\hline
\multicolumn{1}{|l|}{} & Yelp2018 & Gowalla & Amazon-Books \\ \hline
negative weight        & 0.4      & 0.7     & 0.6          \\ \hline
margin                 & 0.9      & 0.9     & 0.4          \\ \hline
regularization         & -9       & -9      & -9           \\ \hline
number of negative     & 800      & 800     & 800          \\ \hline
number of positive     & 10       & 20      & 50           \\ \hline
\end{tabular}
\end{table}
\section{Proof}\label{app:proof}

\subsection{Proof of \cref{th1}}

\begin{proof}
To derive the alternating least square solution, we have: 
\begin{itemize}
    \item Fixing item vectors, optimizing: 
    \begin{equation*}
    \begin{split}
        \mathcal{L}_u &=||\sqrt{c_u}{(H^{u}_S)}^T\mathbf{w}_u - \sqrt{c_u}\mathbf{y}_S||^2 + ||\sqrt{\alpha_0}H^T\mathbf{w}_u||^2 + ||\sqrt{\lambda_u}I \cdot \mathbf{w}_u||^2\\
        &-||\sqrt{c_u\alpha_0} {(H^{u}_S)}^T\mathbf{w}_u||^2, \ \ \ \ where \ \ \ \ \lambda_u = \lambda(|\mathcal{I}_u^+|+\alpha_0|\mathcal{I}|)^{\nu} 
    \end{split}
\end{equation*}
Note that the observed item matrix for user $u$ as $H^u_S$:

\begin{equation}
H^u_S=\begin{bmatrix}
\vrule & \vrule &        & \vrule\\
\mathbf{h}_1 & \mathbf{h}_s & \cdots & \mathbf{h}_{|\mathcal{I}_u^+|}\\
\vrule & \vrule &        & \vrule\\
\end{bmatrix}\in \mathbb{R}^{k\times |\mathcal{I}_u^+|},\quad s\in \mathcal{I}_u^+
\end{equation}

$H$ is the entire item matrix for all items. 
$y_S$ are all $1$ column (the same row dimension as $H^u_S$.

    \item Fixing User Vectors, optimizing: 
 \begin{equation*}
    \begin{split}
        \mathcal{L}_i &=||{(W^{i}_S \sqrt{C_u})}^T\mathbf{h}_i - \sqrt{C_u}\mathbf{y}_S||^2 + ||\sqrt{\alpha_0}W^T\mathbf{h}_i||^2 + ||\sqrt{\lambda_i}I \cdot \mathbf{h}_i||^2\\
        &-||\sqrt{\alpha_0} {(\sqrt{C_u}W^{i}_S)}^T\mathbf{h}_i||^2, \ \ \ \ where \ \ \lambda_i = \lambda(|\mathcal{U}_i^+|+\alpha_0|\mathcal{U}|)^{\nu}
    \end{split}
\end{equation*}
Here, $W^{i}_S$ are observed user matrix for item $i$, and $W$ are the entire user matrix, $C_u=diag(c_u)$ a $|\mathcal{U}|\times|\mathcal{U}|$ diagonal matrix  with $c_u$ on the diagonal.  
\end{itemize}

Solving $\mathcal{L}_u$ and $\mathcal{L}_i$, we have the following closed form solutions: 
\begin{equation*}
    \begin{split}
        & \mathbf{w}_u^*=\Big(c_u (1 -\alpha_0)H^{u}_S{(H^{u}_S)}^T + \alpha_0 HH^T + \lambda_u I\Big)^{-1}\cdot H^{u}_S\cdot \sqrt{c_u}\mathbf{y}_S \\ 
        & \mathbf{h}_i^*=\Big((1 -\alpha_0)W^{i}_SC_u{(W^{i}_S)}^T + \alpha_0 WW^T + \lambda_i I\Big)^{-1}\cdot W^{i}_S\cdot \sqrt{C_u}\mathbf{y}_S
    \end{split}
\end{equation*}

Assuming $c_u$ be a constant for all users, we get 

\begin{equation*}
    \begin{split}
        & \mathbf{w}_u^* \varpropto \Big(H^{u}_S{(H^{u}_S)}^T + \frac{\alpha_0}{(1-\alpha_0)c_u} HH^T + \frac{\lambda_u}{(1-\alpha_0)c_u} I\Big)^{-1}\cdot H^{u}_S\cdot \mathbf{y}_S \\ 
        & \mathbf{h}_i^* \varpropto \Big(W^{i}_S{(W^{i}_S)}^T + \frac{\alpha_0}{(1-\alpha_0)c_u} 
        WW^T + \frac{\lambda_i}{(1-\alpha_0)c_u} I\Big)^{-1}\cdot W^{i}_S \cdot \mathbf{y}_S
    \end{split}
\end{equation*}
Interestingly by choosing the right $\alpha_0$ and $\lambda$, the above solution is in fact the same solution (up to constant factor) for the original $\mathcal{L}_{iALS}$ ~\cite{hu2008collaborative}. Following the above analysis and let 
$\alpha_0^\prime=\frac{\alpha_0}{(1-\alpha_0)c_u}$ and 
$\lambda^\prime=\frac{\lambda}{(1-\alpha_0)c_u}$
\end{proof}

\section{ADMM}
\subsection{ADMM}
ADMM problem: 

$f,g$ convex. ($x,z$ are vectors)

\begin{equation}
    \begin{split}
        &\min f(x) + g(z)\\
        &s.t \quad Ax+Bz = c
    \end{split}
\end{equation}

Its equivalent to optimize:

\begin{equation}
    \begin{split}
        \mathcal{L}_{\rho}(x,y,z)&=f(x)+g(z)\\
        &+y^T(Ax+Bz-c)+(\rho/2)||Ax+Bz-c||^2_2
    \end{split}
\end{equation}

The alternating solution is:

\begin{equation}
\begin{aligned}
x^{k+1} & :=\operatorname{argmin}_x L_\rho\left(x, z^k, y^k\right) \\
z^{k+1} & :=\operatorname{argmin}_z L_\rho\left(x^{k+1}, z, y^k\right) \\
y^{k+1} & :=y^k+\rho\left(A x^{k+1}+B z^{k+1}-c\right)
\end{aligned}
\end{equation}

\subsection{Low-Rank EDLAE}
In \citep{edlae}, equation 10 is to optimize:

\begin{equation}
    \begin{split}
        &||X-X\cdot\{UV^T-dMat\big(diag(UV^T)\big)\}||^2_F\\
        &+||\Lambda^\frac{1}{2}\cdot \{UV^T-dMat\big(diag(UV^T)\big)\}||^2_F
    \end{split}
\end{equation}

This can be re-written as:

\begin{equation}
    \begin{split}
        &||X-XUV^T+X\cdot dMat(\beta)||^2_F+||\Lambda^\frac{1}{2}\cdot (UV^T-dMat(\beta)) ||^2_F\\
        &s.t. \quad \beta = diag(UV^T)
    \end{split}
\end{equation}

Using augmented Lagrangian:

\begin{equation}
    \begin{split}
       & ||X-XUV^T+X\cdot dMat(\beta)||^2_F+||\Lambda^{\frac{1}{2}}UV^T||-||dMat(\beta)||^2_F\\
       &+2\gamma^T\Omega (\beta-diag(UV^T))+||\Omega^\frac{1}{2}(\beta-diag(UV^T))||^2_F
    \end{split}
\end{equation}

$\gamma$ is the vector of Lagrangian multipliers, amd $\Omega = \Lambda + \omega I$ where $\omega>0$ is a scalar.

\begin{equation}
\begin{aligned}
\hat{\mathbf{U}} & \leftarrow\left(\mathbf{X}^{\top} \mathbf{X}+\Lambda+\Omega\right)^{-1}\left(\mathbf{X}^{\top} \mathbf{X} \cdot \operatorname{dMat}(\mathbf{1}+\hat{\beta})+\Omega \cdot \operatorname{dMat}(\hat{\beta}-\hat{\gamma})\right) \hat{\mathbf{V}}\left(\hat{\mathbf{V}}^{\top} \hat{\mathbf{V}}\right)^{-1} \\
\hat{\mathbf{V}}^{\top} & \leftarrow\left(\hat{\mathbf{U}}^{\top}\left(\mathbf{X}^{\top} \mathbf{X}+\Lambda+\Omega\right) \hat{\mathbf{U}}\right)^{-1} \hat{\mathbf{U}}^{\top}\left(\mathbf{X}^{\top} \mathbf{X} \cdot \operatorname{dMat}(\mathbf{1}+\hat{\beta})+\Omega \cdot \operatorname{dMat}(\hat{\beta}-\hat{\gamma})\right) \\
\hat{\beta} & \leftarrow \frac{\operatorname{diag}\left(\mathbf{X}^{\top} \mathbf{X} \hat{\mathbf{U}} \hat{\mathbf{V}}^{\top}\right)-\operatorname{diag}\left(\mathbf{X}^{\top} \mathbf{X}\right)+\operatorname{diag}(\Omega) \odot\left(\operatorname{diag}\left(\hat{\mathbf{U}} \hat{\mathbf{V}}^{\top}\right)+\hat{\gamma}\right)}{\operatorname{diag}\left(\mathbf{X}^{\top} \mathbf{X}\right)+\operatorname{diag}(\Omega-\Lambda)} \\
\hat{\gamma} & \leftarrow \hat{\gamma}+\operatorname{diag}\left(\hat{\mathbf{U}} \hat{\mathbf{V}}^{\top}\right)-\hat{\beta}
\end{aligned}
\end{equation}

\section{EASE-debiased}

\begin{equation*}
    \begin{split}
&\mathcal{L}^{Debiased}_{mse}=\sum\limits_{u} \Bigg[ \sum\limits_{i\in \mathcal{I}_u^+}[c_u(\hat{y}_{ui}-1)^2 -c_u\lambda \hat{y}^2_{ui}]+ \lambda\sum\limits_{t\in\mathcal{I}} \hat{y}_{ut}^2 \Bigg]\\
        &=\sum\limits_{u} \Bigg[ \sum\limits_{i\in \mathcal{I}_u^+}[c_u(\hat{y}_{ui}-1)^2 -c_u\lambda \hat{y}^2_{ui}]+ \lambda\sum\limits_{t\in\mathcal{I}_u^+} \hat{y}_{ut}^2 + \lambda\sum\limits_{p\in\mathcal{I}\backslash \mathcal{I}_u^+} \hat{y}_{ut}^2 \Bigg]\\
&=\sum\limits_{u} \Bigg[\sum\limits_{i\in \mathcal{I}_u^+}[c_u(\hat{y}_{ui}-1)^2+c_u \sum\limits_{p\in\mathcal{I}\backslash \mathcal{I}_u^+} \hat{y}_{ut}^2 \Bigg]\\
&+ \lambda (1-c_u)\sum\limits_{u}\sum\limits_{i\in \mathcal{I}_u^+}\hat{y}^2_{ui} + (\lambda-c_u) \sum\limits_{p\in\mathcal{I}\backslash \mathcal{I}_u^+} \hat{y}_{ut}^2 \\
 &=||\sqrt{C_u}(X-XW)||^2_F \\
        &-\lambda||X\odot \sqrt{C_u-I}XW||^2_F -||(1-X)\odot \sqrt{C_u-\lambda I}XW||^2_F\\
    \end{split}
\end{equation*}

For Regression based $W = UV^T$
\begin{equation*}
\begin{split}
 &L=||\sqrt{C_u}(X-XUV^T)||^2_F \\
        &-\lambda||X\odot \sqrt{C_u-I}XUV^T||^2_F -||(1-X)\odot \sqrt{C_u-\lambda I}XUV^T||^2_F\\
        &+\gamma ||UV||^2_F\\
        &s.t. \quad diag(UV)=0\\
    \end{split}
\end{equation*}

\end{document}